\documentclass{vgtc}                          % final (conference style)
\graphicspath{{figures/}{pictures/}{images/}{./}} % where to search for the images

\usepackage{times}                     % we use Times as the main font
\usepackage{tabu}                      % only used for the table example
\usepackage{booktabs}                  % only used for the table example
\usepackage{lipsum}                    % used to generate placeholder text
\usepackage{mwe}                       % used to generate placeholder figures

\usepackage{mathptmx}                  % use matching math font

\onlineid{4384}

\vgtccategory{Research}

\vgtcinsertpkg

\title{A Simplified Positional Cell Type Visualization using Spatially Aggregated Clusters}

\author{Lee Mason\thanks{e-mail: masonlk@nih.gov}\\ %
       \parbox{1.4in}{\scriptsize \centering National Cancer Institute \\ Queen's University Belfast} %
 \and  Jonas Almeida\\ %
      \scriptsize National Cancer Institute}

\teaser{
  \centering
  \includegraphics[width=.80\linewidth]{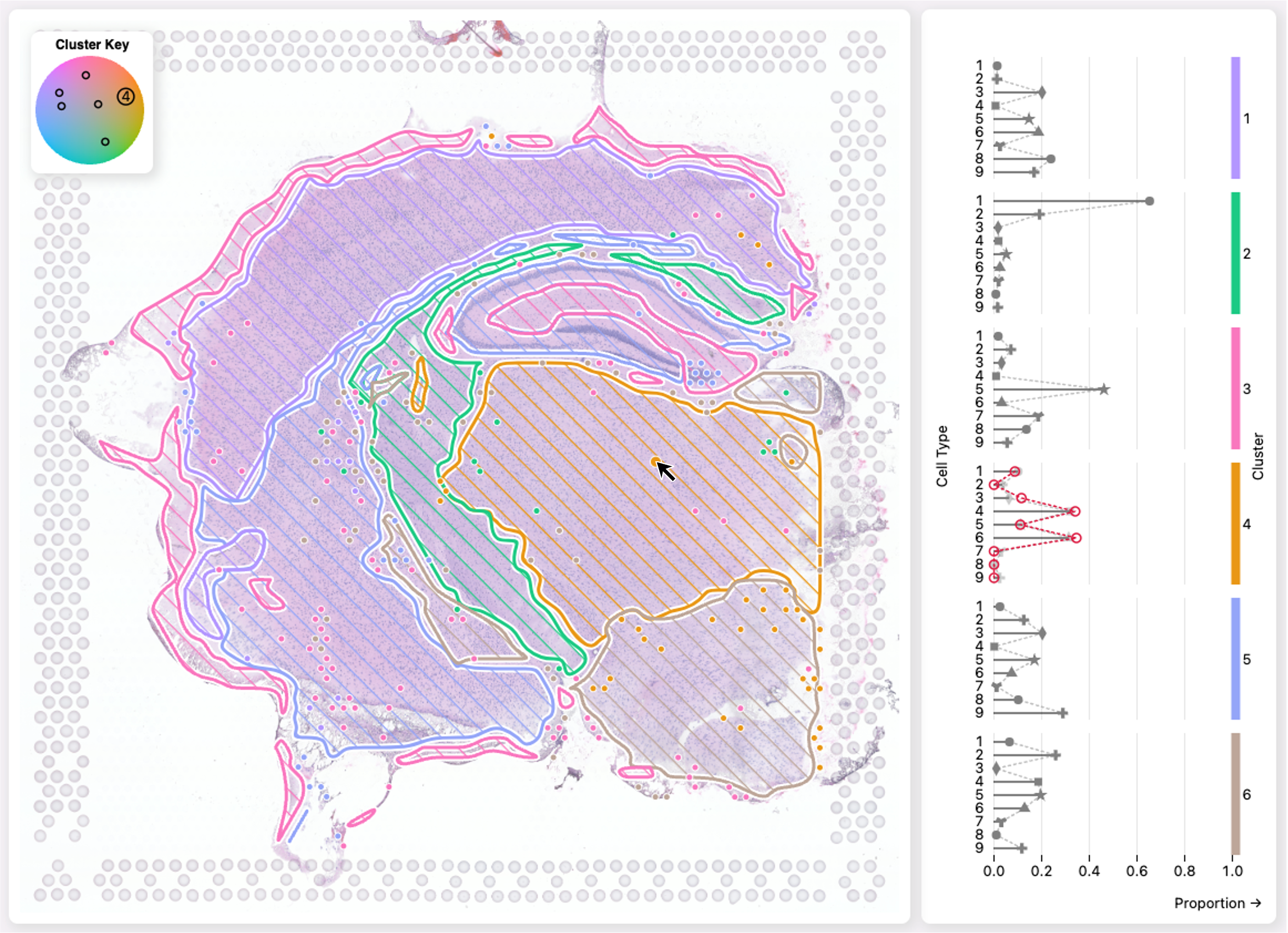}
  \caption{a) A screenshot of the SpotCluster dashboard demonstrating the proposed overlay design (left) and a suggestion of how it can be linked to cluster summaries (right). Here, the user has hovered over a spot on the slide image and that interaction has been reflected in the cluster summary panel. This demo dashboard can be accessed at \url{https://episphere.github.io/spotcluster}.}
  \label{fig:teaser}
}

\abstract{
We introduce a novel method for overlaying cell type proportion data onto tissue images. This approach preserves spatial context while avoiding visual clutter or excessively obscuring the underlying slide. Our proposed technique involves clustering the data and aggregating neighboring points of the same cluster into polygons.
} % end of abstract

\keywords{Clustering, k-means, cluster embedding, visual analytics, interactive dashboard, H\&E, tissue slide image}

\begin{document}


\firstsection{Identifying the problem}

\maketitle

The common approach of using pie charts to visualize spatially positioned cell type proportions \cite{millerReferencefreeCellType2022} presents challenges due to the inherent limitations of pie charts. While some research suggests that pie charts are effective for conveying how proportions relate to the whole, they are generally considered less accurate than other chart types for communicating specific values and comparing proportions \cite{siirtolaCostPieCharts2019}. Additionally, pie charts become increasingly difficult to interpret as the number of categories --- potentially large in the case of cell types ---increases \cite{siirtolaDissectingPieCharts2019}. Comparing pie charts to each other is particularly challenging, even with a small number of charts \cite{kozakMultiplePieCharts2015}. This particular application of pie charts is an example of a location based graph symbol (LGS) approach, which are more commonly applied to geospatial visualization tasks \cite{choengsa-ardEffectiveGraphicFeatures2013}. The challenges remain the same: it is difficult to effectively extract global information from spatially arranged pie charts or even to compare a pie chart to its immediate neighbors. While pie charts are more spatially efficient than linear alternatives, meaning they occlude less of the underlying slide image, occlusion still remains one of the primary limitations of any LGS approach. In this work we will present an alternative that provides simplified, spatially aware global overview of the data while limiting occlusion of the underlying tissue slide.

\section{Our solution: spatially aggregated clustering}

\begin{figure*}[t]
  \centering 
  \includegraphics[width=.85\textwidth, alt={Screenshot}]{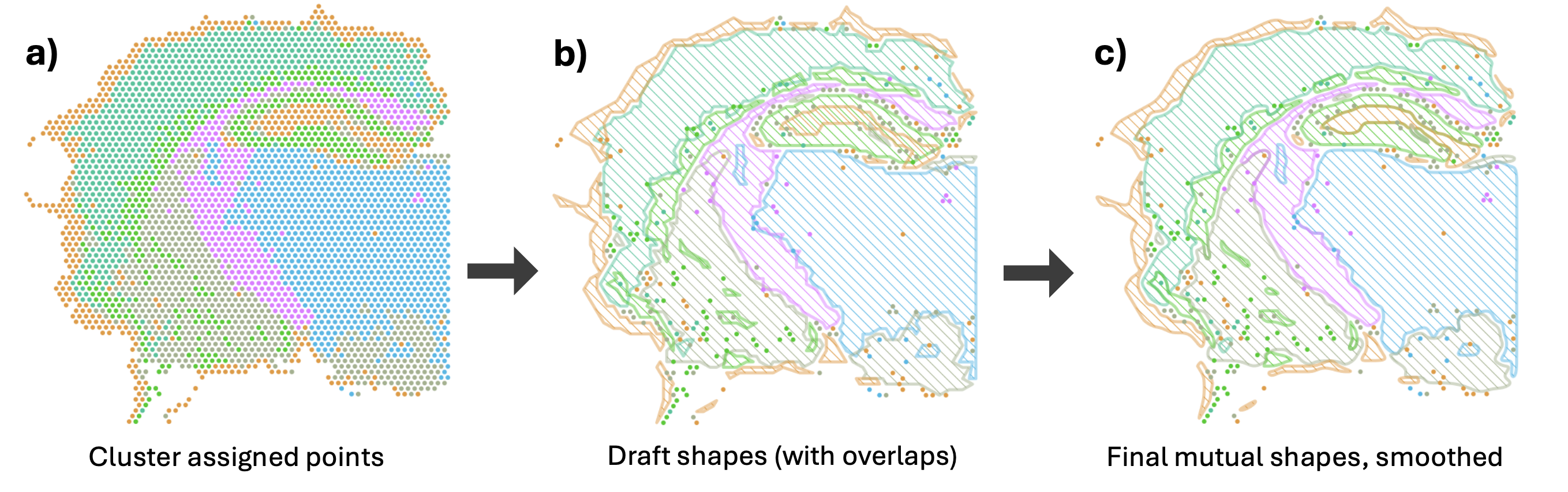}
  \caption{A breakdown of the proposed solution. a) We first cluster the points using k-means, assign a color to each cluster, and color each point according to its assigned cluster label. b) We aggregate neighboring points with the same cluster and generate polygons of each group with a concave hull algorithm. c) We subtract overlapping polygons to ensure polygons are exclusive.}
  \label{fig:breakdown}
\end{figure*}

Our proposed solution leverages our previous work on cluster embedding, where we quantized vector data using partition clustering and summarized the resulting clusters with a lower dimensional representation \cite{masonEpiVECSExploringSpatiotemporal2023}. Our proposed method begins by applying this clustering and dimensionality reduction technique to the spot cell type proportion vectors. Each vector is assigned to a cluster, and the resulting low-dimensional cluster representations are mapped to a perceptually uniform color space (OKHSL). This color mapping ensures that visual similarity between clusters reflects their underlying similarity in the data. Finally, each spot is colored according to its assigned cluster. The issue of points obscuring the image remains, but since points within the same cluster tend to be located near each other on the tissue, we can further simplify the visualization by grouping neighboring points with the same label and replacing the individual points with a representation of the polygon area they form (see \cref{fig:breakdown}b). Creating these polygons is challenging due to the complexity of the shapes formed by the points. In our method, we employ a modern concave hull algorithm implemented in the JavaScript library \texttt{concaveman}. This algorithm effectively encapsulates most points within polygons but it struggles with more intricate concave features. To refine the polygons and ensure they don't overlap, we identify any intersecting shapes and subtract them from each other. The polygon with fewer correctly labeled points within the intersecting area is subtracted from the other. The result is a set of non-intersecting polygons each corresponding to a cluster assignment (see \cref{fig:breakdown}c). We overlay these shapes on the tissue image, coloring each according to its corresponding cluster. This replaces the individual point representations for points within a polygon, but we retain any points which are not inside a polygon or are inside a polygon corresponding to a differing cluster. To distinguish the polygons from the background, and to ensure their colors are clear, we outline them in white. Due to the complex, interwoven concavities present in some of the shapes, it can be difficult to tell which area a shape encapsulates from its outline alone. To address this, we fill each shape with a striped pattern. The striped pattern allows the color to be clearly displayed without occluding the underlying slide image. The resulting overlay provides a simplified summary of the cell types while retaining spatial context, which can act as an informative static summary of the data or provide a home for further interactive exploration. We demo this in a prototype dashboard environment (see \cref{fig:teaser}) where the user can see the encapsulated cluster overview alongside cluster summaries, and interact to view the details of spots and how they fit into their cluster. Each cluster's centroid is displayed using a dot plot, with the dots representing cell type proportions. Dashed lines connect the dots for easier comparison. Dots vary in shape per cell type, aiding visual comparison between clusters. Hovering over a spot on the slide image overlays cell type proportions onto the relevant cluster's dot plot.

\section{Discussion \& Future Work}

While pie charts offer a detailed way to view cell type proportions within spatial context, comparing many pie charts is difficult. We propose a new visualization design that enhances the natural patterns viewers extract from the original design.
At first glance, viewers are likely to notice pie charts that appear similar and are spatially proximate, then naturally group these into contiguous regions. This leverages the Gestalt principles of similarity (noticing spots with similar proportions), proximity (noticing how similar spots tend to be near each other), and continuity (connecting these proportionally and spatially similar spots into regions). Our solution formalizes this intuitive visual analysis pipeline, employing appropriate computational methods to systematically implement each step, and refining the clarity and efficiency of the natural visual comparisons a viewer makes. While detailed information is omitted, it's unlikely that displaying all of this at once would be of much practical value. Experimental evidence suggests that when a viewer is inspecting an LGS visualization, their attention is generally limited to a few neighboring elements, with true focus on just one at a time \cite{choengsa-ardEffectiveGraphicFeatures2013}. Thus, we can hide this information until the user requests it through interaction, sacrificing little of the immediacy of information gained from inspection. In return, we simplify the overall visualization, reduce the viewer's cognitive burden in exploring it, and minimize occlusion of the slide image.

Future work could refine various stages of the pipeline, particularly by systematically comparing different clustering and shape detection methods. The current cluster color assignment method often results in colors too close to the underlying slide image, making shape distinction difficult. While white outlines help, using a color space that avoids the colors of the underlying slide image could provide a superior solution.

\bibliographystyle{abbrv-doi-narrow}

\bibliography{references}
\end{document}